\documentclass[letterpaper, 10 pt, journal, twoside]{IEEEtran}
\ifCLASSINFOpdf
\else
\fi
\newtheorem{definition}{Definition}
\newtheorem{property}{Property}

\usepackage{epsfig}
\usepackage{graphicx}
\usepackage{wrapfig}
\usepackage[flushleft]{threeparttable}
\usepackage{footnote}

\usepackage{graphics} 
\usepackage{graphicx}
\usepackage{grffile} 
\usepackage{amsmath} 
\usepackage{amssymb}  

\usepackage{pifont}
\usepackage{color}
\usepackage{cite}
\usepackage{bm}
\usepackage{url}

\usepackage[noend]{algpseudocode}
\usepackage{algorithm}
\usepackage{makecell}
\usepackage{multirow}

\usepackage{enumitem}

\usepackage{booktabs}

\usepackage[pagebackref=true,breaklinks=true,colorlinks,bookmarks=false]{hyperref}
\hypersetup{colorlinks, citecolor={black}, linkcolor={black}}

\usepackage{stfloats}

\newcommand{\rebuttal}[1]{\textcolor{black}{ #1}}

\definecolor{darkviolet}{rgb}{0.58, 0.0, 0.83}

\DeclareMathOperator*{\argmax}{arg\,max}
\DeclareMathOperator*{\argmin}{arg\,min}

\makeatletter
\DeclareRobustCommand{\qed}{%
  \ifmmode 
  \else \leavevmode\unskip\penalty9999 \hbox{}\nobreak\hfill
  \fi
  \quad\hbox{\qedsymbol}}
\newcommand{\openbox}{\leavevmode
  \hbox to.77778em{%
  \hfil\vrule
  \vbox to.675em{\hrule width.6em\vfil\hrule}%
  \vrule\hfil}}
\newcommand{\qedsymbol}{\openbox}
\newenvironment{proof}[1][\proofname]{\par
  \normalfont
  \topsep6\p@\@plus6\p@ \trivlist
  \item[\hskip\labelsep\itshape
    #1.]\ignorespaces
}{%
  \qed\endtrivlist
}
\newcommand{\proofname}{Proof}
\makeatother

\begin{document}
%

\title{
Concavity-Induced Distance \\ for Unoriented Point Cloud Decomposition
}

\author{Ruoyu Wang$^{1}$, Yanfei Xue$^{1}$, Bharath Surianarayanan$^{1}$, Dong Tian$^{2}$, and Chen Feng\textsuperscript{1, \ding{41}} 

\thanks{Manuscript received: February, 8, 2023; Revised May, 6, 2023; Accepted June, 1, 2023.}
\thanks{This paper was recommended for publication by Editor Markus Vincze upon evaluation of the Associate Editor and Reviewers' comments.
This work was mainly supported by InterDigital Inc. in 2021, and partly by NSF grant DUE-2026479 in 2022.} 
\thanks{$^{1}$Ruoyu Wang, Yanfei Xue, Bharath Surianarayanan, and Chen Feng are with Tandon School of Engineering, New York University, United States. Chen Feng is the corresponding author
        {\tt\footnotesize \{ruoyuwang, yx2066, bs4224, cfeng\}@nyu.edu}}%
\thanks{$^{2} $Dong Tian is with InterDigital Inc., United States
        {\tt\footnotesize Dong.Tian@interdigital.com}}%
\thanks{Digital Object Identifier (DOI): see top of this page.}}
\markboth{IEEE Robotics and Automation Letters. Preprint Version. Accepted June, 2023}
{Wang \MakeLowercase{\textit{et al.}}: CONCAVITY-INDUCED DISTANCE} 

%



\maketitle

\begin{abstract}
We propose Concavity-induced Distance (CID) as a novel way to measure the dissimilarity between a pair of points in an unoriented point cloud. CID indicates the likelihood of two points or two sets of points belonging to different convex parts of an underlying shape represented as a point cloud. After analyzing its properties, we demonstrate how CID can benefit point cloud analysis without the need for meshing or normal estimation, which is beneficial for robotics applications when dealing with raw point cloud observations. By randomly selecting very few points for manual labeling,
a CID-based point cloud instance segmentation via label propagation achieves comparable average precision as recent supervised deep learning approaches, on S3DIS and ScanNet datasets. Moreover, CID can be used to group points into approximately convex parts whose convex hulls can be used as compact scene representations in robotics, and it outperforms the baseline method in terms of grouping quality. \rebuttal{Our project website is available at: https://ai4ce.github.io/CID/}
\end{abstract}

\begin{IEEEkeywords}
Object Detection, Segmentation and Categorization, Computational Geometry
\end{IEEEkeywords}
\section{INTRODUCTION}
Convexity-based shape analysis has been widely used in many robotics tasks. For instance, for collision detection in path planning~\cite{schulman2013finding}, non-convex-shaped obstacles need to be decomposed into convex ones to accelerate the computation. Another example is shape segmentation in 3D scene understanding, where convex partitioning of objects is shown to be useful in robotics~\cite{stein2014convexity}. It has also been used to help with object grasping~\cite{chari2012convex} and human gesture recognition~\cite{qin2014real}.

In all those tasks, meshes~\cite{ghosh2013fast} or volumetric 3D models~\cite{mamou2016volumetric} of objects or scenes are standard input to convexity-based shape analysis approaches such as approximate convex decomposition (ACD)~\cite{lien2004approximate}. However, few of those approaches can be directly applied to process unoriented point clouds that are widely used in robotics. And it is non-trivial to obtain meshes or volumetric models from unoriented point clouds that are directly captured by LiDAR or 3D cameras, which typically involves time-consuming post-processing steps like normal estimation, normal direction alignment, and surface reconstruction.

\begin{figure}[t!]
    \centering
    \includegraphics[width=0.45\textwidth]{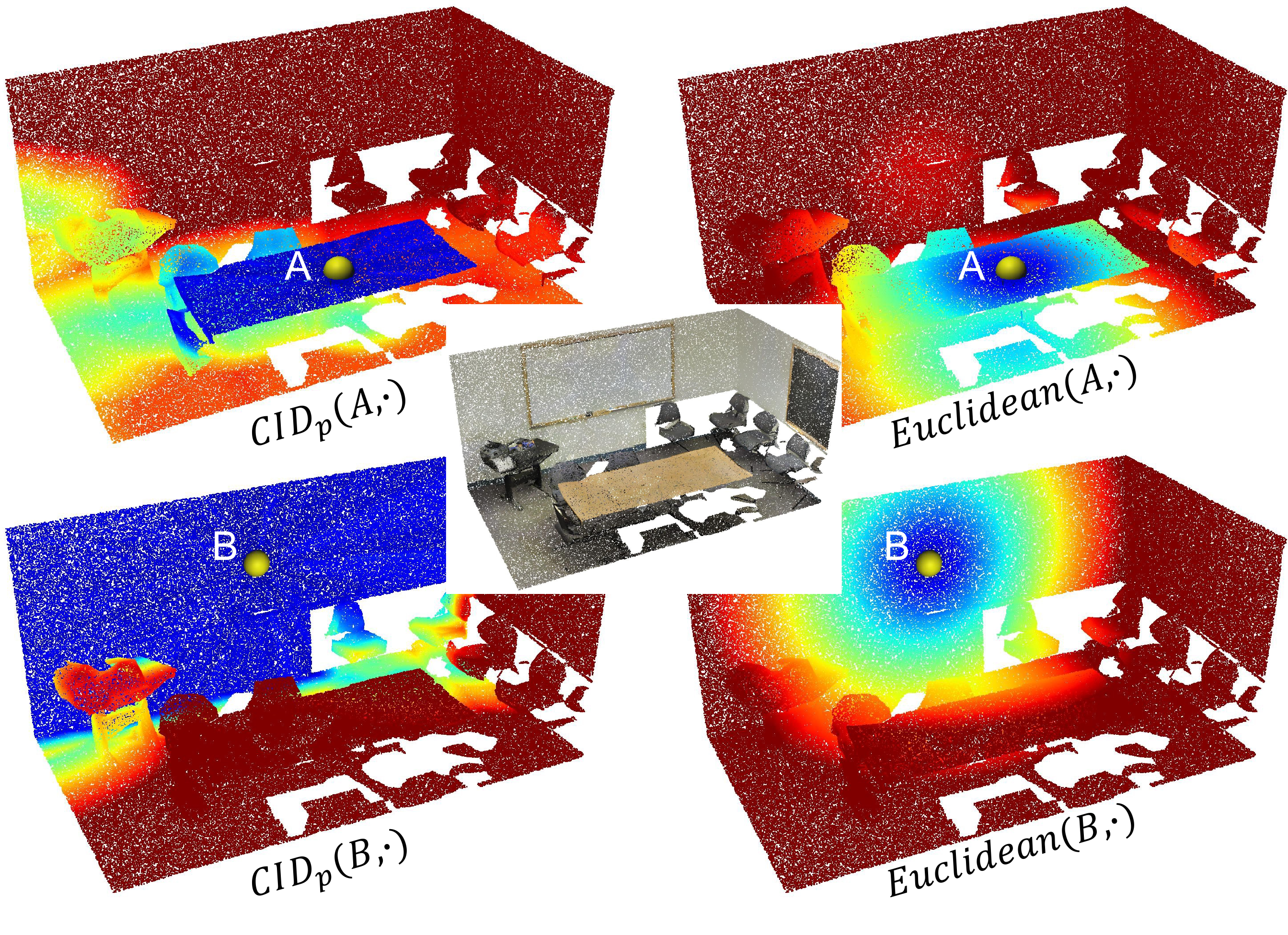}
    \vspace{-3mm}
    \caption{\textbf{Visualization of CID vs. Euclidean distance.} A is a point on the table and B is a point on the wall.  The images on the left (or right) column show the CID (or the Euclidean distance) from A or B to each point in the scene (shown as the center image). Blue/Red means a smaller/larger distance. This shows the stronger discriminative power of CID than Euclidean distance.}
    \label{fig:vis_cid}
    \vspace{-5mm}
\end{figure}

Therefore, we are motivated to develop a convexity-based shape analysis approach that is directly applicable to unoriented point clouds. This requires us to first define shape convexity on point clouds, which brings two challenges:

\textit{Defining convexity of a discrete set.}
A convex set requires that each line segment joining every two points in this set is still within this set. However, a point cloud is the discretization of a continuous shape surface, i.e., a set of points sampled from the surface.
Therefore, the conventional convexity definition cannot be directly used on a point cloud to evaluate the convexity of the underlying shape. Note that although the volumetric 3D model is also a discretized shape representation, its voxels are organized and have volume, making the conventional definition still applicable.

\textit{Discovering surface orientation via point clouds.} To address the above challenge, some convexity-based shape analysis approaches are designed for point cloud sampled from CAD models~\cite{asafi2013weak, kaick2014shape} or captured by RGBD cameras~\cite{christoph2014object, gong2017point, stein2014convexity}. To evaluate shape convexity, both cases leverage the information about surface orientation such as oriented normals, which is trivially obtained because these point clouds are naturally oriented. Therefore, these approaches still cannot be directly applied to unoriented point clouds.

We address these two challenges in this paper. For the first one, we can define a function to capture the ``concavity'', i.e., how likely a line segment joining any two points on the point cloud is outside of the underlying shape. This can be evaluated by the distance from the line segment to the point cloud. Intuitively, the large that distance is, the more likely the line segment is outside the underlying shape. For the second challenge, we propose to simply ignore the surface orientation during the analysis. This means that even if the above line segment lies inside the underlying shape, the above function is still going to output a large ``concavity'' value, as long as it is far from the surface. 

This leads us to define the Concavity-Induced Distance (CID) between two points or two sets of points (see Figure~\ref{fig:vis_cid}). CID does not require surface orientation and can be calculated on a set of discretized points coordinates without any additional information. We show that the CID between two points exhibits some ideal properties such as rotational and translational invariance, which is useful for measuring point similarities for semantic or instance segmentation tasks. Figure~\ref{fig:vis_cid} visualizes the CIDs and the Euclidean distances between a selected point and all other points in the point cloud. We can see that points on the same convex partition (on the same wall, on the surface of the table) as the selected point have smaller CID than those on the different convex partitions. Therefore, CID has the potential to separate points from different convex partitions in an unoriented point cloud. Considering the boundaries between convex partitions usually align with the boundaries between objects or object parts, this makes CID useful for segmentation-based scene understanding.

In this work, we show the effectiveness of CID by applying it to two scene understanding tasks: instance segmentation and scene abstraction on unoriented point clouds. Nowadays, such tasks are usually addressed by deep learning approaches, which typically requires large-scale manually labeled datasets, especially for segmentation with point-wise labeling. For example, the widely used S3DIS~\cite{2017arXiv170201105A} dataset contains over 695 million labeled points.
\textit{As a complementary and orthogonal approach, using CID to segment unoriented point clouds could make such manual labeling much more efficient}.
In summary, our contributions are listed below:
\begin{itemize}
  \item We propose concavity-induced distance (CID), a novel way to measure the surface concavity between two points or two point sets in an unoriented point cloud.
  \item We show a CID-based label propagation for point cloud instance segmentation on unoriented scene-level point clouds, which achieves comparable performance to recent supervised deep learning methods, and thus can be used to improve point-wise labeling efficiency.
  \item We show a CID-based scene abstraction, which can identify nearly convex parts in an unoriented point cloud. The abstracted scene is useful for robotics tasks like collision detection in path planning.
\end{itemize}
\section{Related Works}
\textbf{Convex shape decomposition.}
The idea of decomposing an arbitrary shape into a set of convex or nearly convex partitions has a long research history and is useful in many fields.
Most of the existing approaches can be categorized into two classes:
The first class of approaches does not directly evaluate the concavity between two points on the shape~\cite{kreavoy2007model, lien2007approximate, lien2008approximate, sheffer2007shuffler, mamou2016volumetric}. 
For example,~\cite{kreavoy2007model} defines the part convexity over a part of a shape, based on its distance to its convex hull. 
The other class of approaches, including ours, requires evaluation of the concavity between two points on the shape~\cite{ghosh2013fast, liu2010convex}. 
For example, \cite{liu2010convex} computes this concavity based on the Reeb graph~\cite{shinagawa1991surface}.
However, these approaches require an organized data format with oriented normals, i.e., the mesh or the volumetric model, while our approach works on unoriented point clouds directly.
The most related work to our approach is \cite{ghosh2013fast} since their definition of concavity measurement has a similar formulation as ours. 
However, their concavity is defined over two vertices on a mesh, and it requires the connectivity information (edges) between the vertices, just like other mesh-based approaches.

\textbf{Convexity-based point cloud segmentation.} 
The idea of using convexity or concave boundaries for point cloud segmentation has been investigated in both vision and robotics.
Some of the approaches take the point clouds sampled from CAD models~\cite{asafi2013weak, kaick2014shape}, and others take the point clouds captured by the RGBD sensors~\cite{christoph2014object, gong2017point, stein2014convexity, tateno2015real}. Again, the oriented normals of point clouds are needed by these approaches. Differently, our CID-based approach does not require the input point cloud to have oriented normals.

\textbf{Convexity in 3D deep learning.}
Recently, with the rise of 3D deep learning, there are some approaches that introduce convexity into 3D deep learning. Cvxnet~\cite{deng2020cvxnet} proposes to learn to reconstruct 3D meshes with a set of convex primitives. However, training the Cvxnet requires the ground truth Signed Distance Function (SDF) for each 3D shape, while the SDF is not available for unoriented point clouds. Besides, the number of convex partitions for the Cvxnet is fixed, while our CID-based segmentation allows a variable number of segments via a merging step. \cite{gadelha2020label} uses V-HACD to provide the self-supervision signal to achieve label-efficient learning for point cloud segmentation. However, V-HACD requires volumetric models, which are not directly applicable to unoriented point clouds.

\textbf{Learning-based compact 3D representation.} 
Representing complex 3D objects or scenes as compact and usually convex geometric primitives is appealing for many tasks.
\cite{tulsiani2017learning} proposes to learn a set of oriented boxes to represent a 3D shape. \cite{zou20173d} learns to generate 3D shapes represented by a sequence of oriented boxes with recurrent neural networks (RNN). Similar to Cvxnet, these approaches also require the SDF of the 3D shape. Differently,  without learning, CID can be used to obtain convex hulls of convex parts as compact 3D representations of unoriented point clouds of large scenes.

\section{Concavity-Induced Distance}

Next, we will first introduce our definition of Concavity-Induced Distance (CID) between two points and two sets of points. Then we will discuss the properties of CID. 

\rebuttal{
Note that the meaning of \textit{concavity} could be somewhat confusing in geometry processing and optimization fields. The word concavity in CID originates from ``concave polygon'' which means \textit{non-convex} shapes, following the convention in previous convexity-based shape analysis works. This is different from the meaning of a ``concave function'' which means the \textit{negative of a convex function whose shape could still be convex}.}

\subsection{CID between two points}  

\begin{definition}[$CID_p$] The $CID_p$ points $p_i, p_j \in S$, given a surface $S \subseteq \mathbb{R}^D $, \rebuttal{which can be a single object or a scene with multiple objects}, is defined as the maximum distance from any point on line segment $\overline{p_ip_j}$ to $S$:
\begin{equation}
    CID_p(p_i, p_j | S) = \max_{p\in \overline{p_i p_j}}d(p; S).
    \label{equ:cidp}
\end{equation}
\label{def:cidp}
\vspace{-5mm}
\end{definition}

The intuition of definition~\ref{def:cidp} comes from the definition of the \textit{mutex pair}~\cite{liu2010convex} in conventional convex shape decomposition: $\forall p_i, p_j\in S$, if $\exists p \in \overline{p_ip_j}, p \not \in S$, then $p_i$ and $p_j$ is called a mutex pair, which means $p_i$ and $p_j$ are not in the same convex part. \textit{However, the mutex pair definition does not work} when $S$ is represented as a countable point set, i.e., point cloud, instead of a continuous surface, because $\forall p_i, p_j\in S$, $\exists p \in \overline{p_ip_j}, p\not \in S$ always holds, because of the ``sampling gaps'' on the object surface. This limits the applicability of the mutex pair in point cloud-related problems. Note that $CID_p(p_i, p_j | S)$ is also  equivalent to the Hausdorff distance between $\overline{p_ip_j}$ and $S$.

Therefore, instead of predicting whether $p_i$ and $p_j$ is a mutex pair, $CID_p(p_i, p_j)$ is  used to measure how likely $p_i$ and $p_j$ is a mutex pair. Here, $d(p;S)$ is the point to set the distance between $p$ and $S$. 
A higher $d(p;S)$ means a lower likelihood that $p \in S$. Therefore, $CID_p$, the maximum $d(p;S)$ for $p \in \overline{p_ip_j}$, could be used to quantify the likelihood that $p_i$ and $p_j$ are from the same convex partition of $S$, which is different from the original definition of mutex pair that checks the \textit{existence} of $p$.

\textbf{Approximation of $CID_p$.} In practice, $S$ is usually represented as a point cloud with $N$ points: $S = \{p_k | k \in [0,N)\}$. Therefore, $d(p; S)$ can be calculated as:
\begin{equation}
    d(p; S)=\min_{0\leq k < N}\|p - p_k\|. \nonumber
\end{equation}
 Besides, $\overline{p_ip_j}$ is discretized into a set of $M$ points $L=\{p_l |  l \in [0,M), p_l\in\overline{p_ip_j}\}$ . The discretization makes the calculation easier to be implemented and parallelized. The $CID_p(p_i, p_j)$ can be approximated as:
\begin{equation}
    CID_p(p_i, p_j | S) \approx \max_{0\leq l < M}\min_{0\leq k < N}\|p_l - p_k\|.
    \label{equ:acidp}
\end{equation}
Equation~\eqref{equ:acidp} is used throughout our experiments.

\begin{figure}[!t]
  \centering
  \hspace{-3mm}\includegraphics[width=0.35\textwidth]{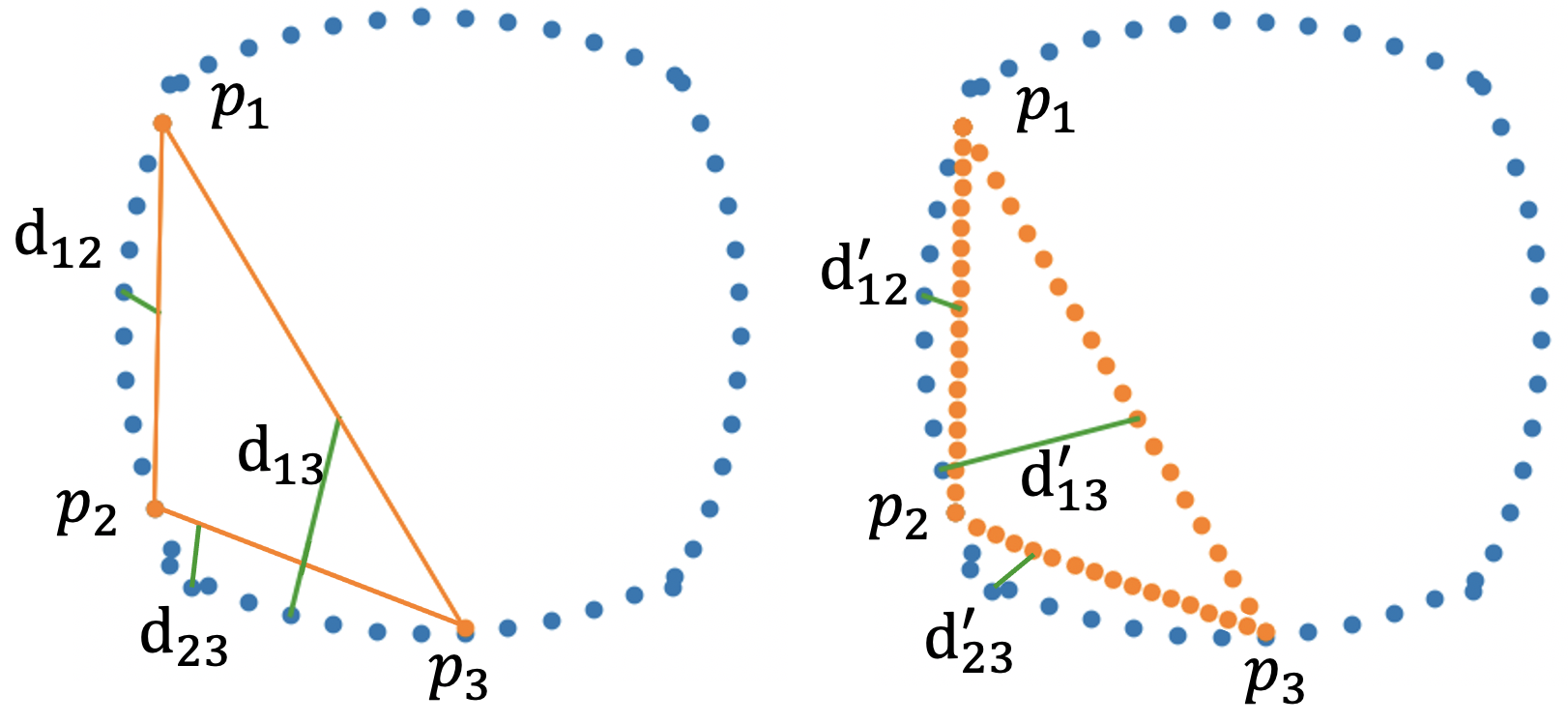}
  \includegraphics[width=0.35\textwidth]{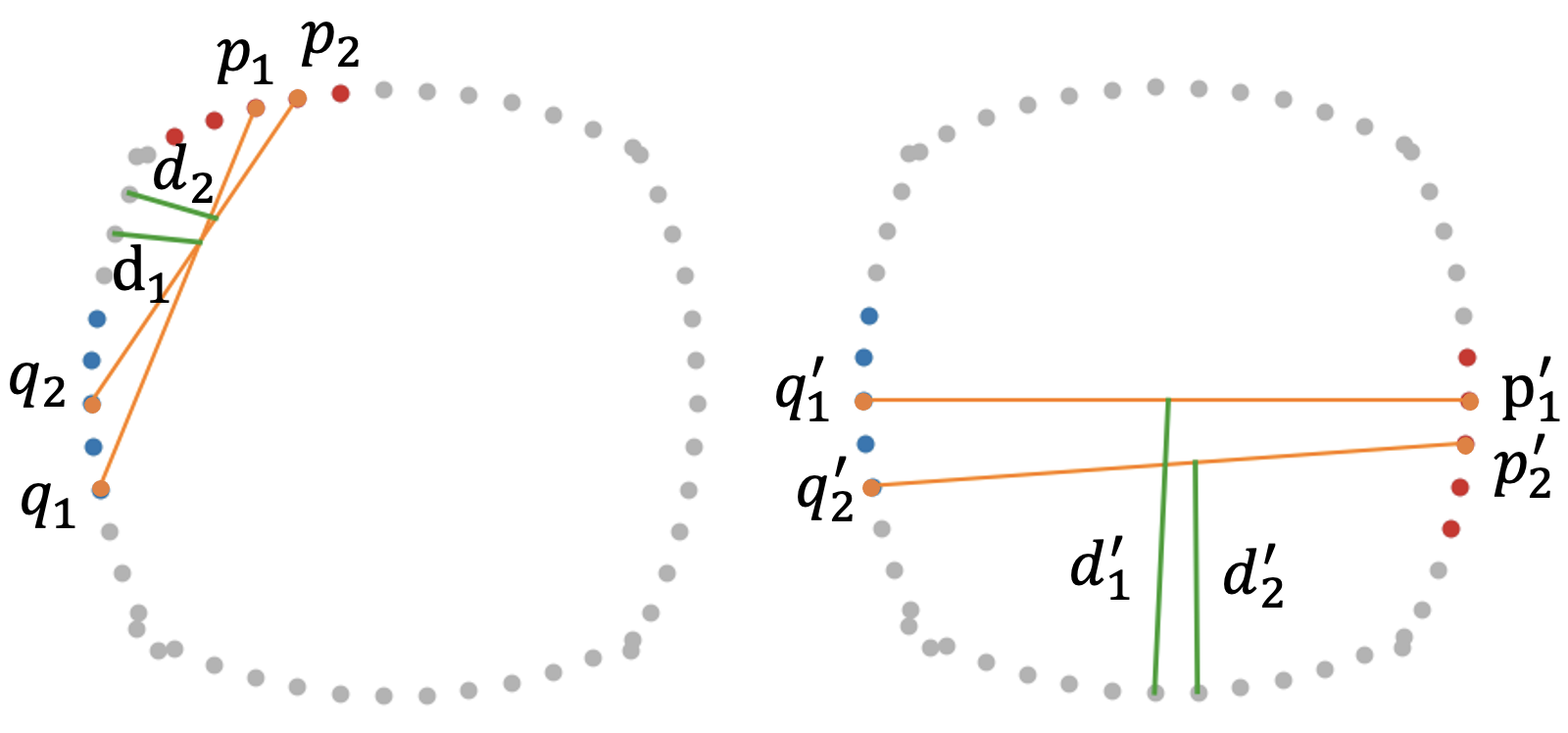}
  \vspace{-2mm}
  \caption{\textbf{$CID$ and its approximation.}
  The top row shows the calculation of $CID_p$ in a four-arc-shaped point cloud. The bottom row  illustrates that of $CID_g$.
  In the top row, the yellow dots $p_1$, $p_2$, and $p_3$ are three points selected. Top left: $d_{12}$, $d_{13}$ and  $d_{23}$ show the $CID_p(p_1, p2)$, $CID_p(p_1, p3)$ and $CID_p(p_2, p3)$ calculated by equation~\ref{equ:cidp}. Top right:  $d'_{12}$,  $d'_{13}, d'_{23}$ show the approximation of $CID_p$ by discretizing the line segments $\overline{p_1p_2}$, $\overline{p_1p_3}$ and $\overline{p_2p_3}$ (equation~\ref{equ:acidp}).
  In the bottom row, the red dots ($G_i$) and blue dots ($G_j$) are two subsets of the point cloud ($S$). $(p_m, q_m)$ and $(p'_m, q'_m)$ ($m=1, 2$) are example point pairs between $G_i$ and $G_j$. Bottom left: smaller $CID_g(G_i, G_j | S)$. Bottom right: larger $CID_g(G_i, G_j | S)$. Best viewed in color.}
  \label{fig:cidp}
  \vspace{-5mm}
\end{figure}

Figure~\ref{fig:cidp} demonstrates $CID_p$ and its approximation. In the figure, it is clear that $d_{13} > d_{12}$, which means that $p_1$ and $p_3$ are more likely to be a mutex pair than $p_1$ and $p_2$. This result also aligns with human intuition since the points between $p_1$ and $p_3$ are more concavely aligned than those between  $p_1$ and $p_2$. Besides, the approximation ($d'_{12}$ and $d'_{13}$) is very close to the accurate $CID_p$ ($d_{12}$ and $d_{13}$).

\subsection{CID between two groups of points}  
\begin{definition}[$CID_g$] The $CID_g$ between two groups of points $G_i, G_j \subseteq  S$, given a point set $S \subseteq \mathbb{R}^D $ is defined as the average $CID_p$ for all pairs of points $(p, q)$, where $p \in G_i, q \in G_j$, and $n(\cdot)$ is the number of points in a point set:

\vspace{-3mm}
\begin{equation}
    CID_g(G_i, G_j | S) = \frac{\sum_{p \in G_i}\sum_{q \in G_j}CID_p(p, q | S)}{n(G_i)n(G_j)}.
    \label{equ:cidg}
\end{equation} 
\label{def:cidg}
\end{definition}
\vspace{-1.5mm}

The definition of $CID_g$ is a very natural extension of $CID_p$. $CID_g$ captures the likelihood that two sets of points are in the same convex partition. Higher $CID_g$ indicates lower likelihood that two sets of points are in the same convex partition.

\textbf{Approximation of $CID_g$.} To improve the computational efficiency, we use uniformly downsampled point set $G'_i\subset G_i, G'_j\subset G_j$ to compute $CID_g(G'_i, G'_j|S)$ as an approximation to $CID_g(G_i, G_j|S)$:
\begin{equation}
    CID_g(G_i, G_j|S) \approx CID_g(G'_i, G'_j|S)
\end{equation}

In Figure~\ref{fig:cidp}, $d_1$ and $d_2$ (left) are smaller than $d'_1$ and $d'_2$ (right), which indicates that point pairs between $G_i$ and $G_j$ usually have smaller $CID_p$ in the left than in the right. Therefore, $CID_g(G_i, G_j | S)$ is smaller on the left than on the right, which means that $G_i$ and $G_j$ on the left are more likely to be in the same convex partition.

\subsection{Properties of CID}
\begin{property}
$CID_p$ is non-negative, symmetric, and reflexive:
\vspace{-2mm}
\begin{align}
&CID(p_i, p_j | S) \geq 0, \nonumber \\ \nonumber
&CID(p_i, p_j | S) = CID(p_j, p_i | S), \\ \nonumber
&CID(p_i, p_i | S) = 0. \nonumber
\end{align}
\label{lem:non-neg}
\vspace{-5mm}
\end{property}

Property~\ref{lem:non-neg} is obvious according to the definition~\ref{def:cidp}. The detailed proof is on our project website.
\begin{property}
$CID_p$ does not satisfy the triangle inequality.\label{lem:tri}
\end{property}
Property~\ref{lem:tri} can be illustrated by the counterexample in Figure~\ref{fig:cidp}. Obviously, $CID_p(p_1, p_3)$ 
$> CID_p(p_1, p_2) + CID_p(p_2, p_3)$. Therefore, the triangle inequality does not hold for $CID_p$. The detailed proof is shown on our project website. 

\textit{Note that not satisfying triangle inequality could be a desirable property for separating objects.} Suppose $p_1, p_2, p_3 \in S$, and $S_1, S_2 \subseteq S$ are two objects with slight overlapping in the point cloud, i.e., $p_1, p_2 \in S_1$, $p_2, p_3 \in S_2$, $p_1 \notin S_2$, $p_3 \notin S_1$. In this case, $p_2$ is on the boundary between the two objects (e.g., the intersection between two walls). A segmentation-friendly distance $D$ on points should output always less than a threshold for any two points on the same object: $D(p_1, p_2)<\epsilon,\, D(p_2, p_3)< \epsilon$. When using $D$ to combine points into the same segment, the smaller the $\epsilon$ we can pick, the better the $D$ is for segmentation.
Now, if $D$ satisfies triangle inequality, just like Euclidean distance, then $D(p_1, p_3) < D(p_1, p_2) + D(p_2, p_3) < 2\epsilon$. But this creates a dilemma. The upper bound of $D(p_1, p_3)$ is $2\epsilon$ which gets smaller at the same speed as the threshold $\epsilon$ used to combine points on the same segment, which makes it harder to separate $p_1$ and $p_3$ when scanning noise and sampling density affects the distance calculation. Without the restriction of the triangle inequality, $CID(p_1, p_3)$ can get arbitrarily large and is not bound by the largest intra-segment CID value, making it easier to separate $p_1$ and $p_3$. 
\begin{property}
$CID_p$ is invariant to rotation and translation.
\end{property}
\begin{proof}
According to equation~\ref{equ:acidp}, $CID_p$ is aggregated from $L$-2 norms, $d(p; S)$, that is rotational and translational invariant, which is preserved under the max operator.
\end{proof}

\textbf{Time and space complexity.} According to equation~\ref{equ:acidp}. The time complexity to compute $CID_p$ is $O(MN)$. In practice, we usually set $M$ constant. In this case, the time complexity becomes $O(N)$. The space complexity is $O(1)$.

\section{Application of CID}

Next, we demonstrate that CID can be used in two important tasks for robotics. One is point cloud instance segmentation and the other is scene abstraction. In the instance segmentation, a small portion of points called seed points need to be proposed and labeled. Then the instance segmentation is performed by propagating the labels to the unlabeled points based on the nearest CID neighbor. In the abstraction task, no label is needed. A point cloud is decomposed into some approximate convex parts based on the CID, and then the convex hull for each part is calculated to abstract the point cloud into a set of convex hulls.
\label{sec:app}
\begin{figure}
    \centering
    \parbox{3in}{\centering\includegraphics[width=0.35\textwidth]{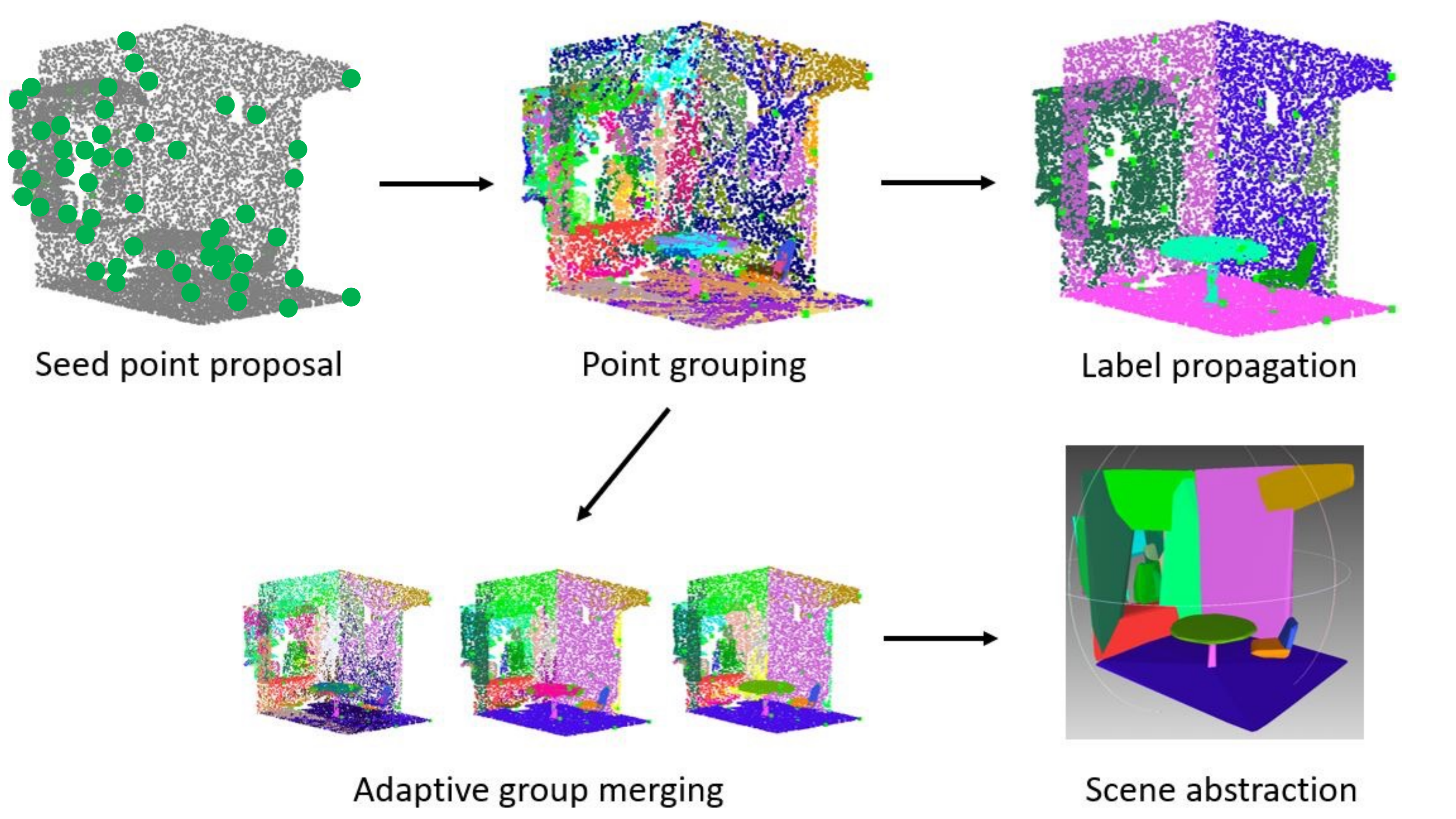}}
    \caption{\textbf{Point cloud instance segmentation (top branch) and point cloud abstraction (bottom branch).} Green dots are proposed seed points. Different groups of points are color-coded.}
    \label{fig:flowchart}
    \vspace{-7mm}
\end{figure}
\subsection{Point cloud instance segmentation via label propagation}
\label{sec:labelprop}

\textbf{Problem definition.} Label propagation is a semi-supervised machine-learning technique that propagates labels from a small set of labeled data points to unlabeled ones based on some rules. 
$CID_p$ can be used to define such rules in label propagation for point cloud instance segmentation. Suppose $S \subset \mathbb{R}^D$ is a point set. $S_l \subset S$ is a seed point set with $K$ points that needs to be labeled. The complement set $S_u \triangleq S \setminus S_l$  is an unlabeled point set with $N$ points. For any point $p_i \in S_l$ with a label $\rho_i$, a group of unlabeled points $G_i$ can be assigned with the same label if $p_i$ is their closest point in $S_l$, in terms of $CID_p$, i.e.,
\begin{equation}
G_i = \{q \ |\  p_i = \argmin_{p\in S_l}CID_p(p, q), q \in S_u\}, \nonumber
\end{equation}
and $S_u = \bigcup_{i=1}^{K} G_{i}$ and  $G_i \cap G_j = \varnothing $, if $i \neq j$. The whole process can be divided into three steps: \textit{seed-point proposal}, \textit{point grouping}, and \textit{group labeling}.

\textbf{Seed-point proposal.} The first step should be selecting seed points that constitute $S_l$. We found that the seed-point proposal method has a significant influence on the performance of label propagation. 
We chose to use CID-based Farthest Point Sampling (CID-FPS) to propose seed points since we want the seed points to be well-distributed into different convex partitions of $S$. \rebuttal{The process of CID-FPS is similar to the original FPS proposed by \cite{qi2017pointnet++}}, while the only difference is that we replace the Euclidean distance with $CID_p$. More details are demonstrated in Algorithm~\ref{alg:spp}. $K$ is a hyper-parameter that determines the number of seed points. Usually, $K$ should be no less than the number of convex partitions in $S$. More complex scenes usually need a larger $K$.
In Figure~\ref{fig:flowchart}, the green dots indicate the proposed seed points.

\begin{algorithm}
\caption{Seed-point proposal by CID-FPS.\label{alg:spp}}
\textbf{Input:} point cloud $S$, number of seed point $K$\\
\textbf{Output:} seed points $S_l\subset S$, remaining points $S_u\subset S$, CID matrix $D$
\begin{algorithmic}
\State randomly pick a point $p^* \in S$ \Comment{\textit{a random start}}
\State $S_l \gets \{p^*\},\quad S_u \gets S \setminus \{p^*\}$
\While{$|S_l| \leq K$}
\ForAll{$p \in S_u$, $q \in S_l$}
        \State $D[p,q] \gets CID_p(p,q)$ \Comment{\textit{cache the $CID_p$ matrix}
    \EndFor}
    \State $D_{\min}[p]\gets \underset{q\in S_l}{\min}\ D[p, q]$ 
    \State $p*\gets \underset{p\in S_u}{\argmax}\ D_{\min}[p]$
    \State $S_l \gets S_l \cup \{p^*\},\quad S_u \gets S_u \setminus \{p^*\}$
\EndWhile
\end{algorithmic}
\end{algorithm}

\textbf{Point grouping.} Once $K$ seed points are proposed, the final $N \times K$ CID matrix $D$ is calculated, as shown in Algorithm~\ref{alg:spp}. $D$ contains $CID_p$ between all pairs of labeled and unlabeled points $(p, q)$, wherein $p\in S_l, q\in S_u$. Each point in $S_l$ will be labeled. Each $q$ can be then assigned to its  $CID_p$-closest seed point $p$, which can be implemented as a row-wise $argmin$ in $D$. Therefore, all points in $S$ can be segmented into $K$ groups.  In Figure~\ref{fig:flowchart}'s point grouping step, different groups are color-coded with different colors.

\textbf{Group labeling.} After grouping, each group of points $G_i$ is assigned with the same label $\rho_i$ from its corresponding seed point.  In Figure~\ref{fig:flowchart}'s label propagation step, points are color-coded by their propagated labels.

\label{sec:propagation}

\subsection{Convexity-based point cloud abstraction}

\textbf{Problem definition.} When labels are not provided, a point cloud can also be decomposed into several approximately convex partitions based on CID. The boundaries between the convex partitions are highly correlated with the boundaries between the object instances. Then the point cloud can be abstracted by a set of convex hulls of all partitions.

The first two steps of point cloud abstraction are the same as  section~\ref{sec:propagation} (seed-point proposal and point grouping). Because there is no ground truth label in the point cloud abstraction task, after point grouping, the initial $K$ ($K$ is also the number of seed points) groups $\{G_i|i\in[0,K)\}$ will be adaptively merged to $K'$ ($K' \leq K$) new groups $\{G'_j\}$ to alleviate over-segmentation.

\textbf{Group merging.} We use a similar iterative greedy merging strategy as in \cite{feng2014fast,kaick2014shape}, as shown in Algorithm~\ref{alg:merge}. For each merging iteration, we calculate $CID_g(G_i, G_j | S)$ for each pair of $(G_i, G_j)$, wherein $i \neq j$.
Then we merge the pair with the lowest $CID_g$ among all pairs. Then the $CID_g$ between each pair of the new groups is recalculated and the merging step is repeated until the number of steps reaches $T$. $T$ is a hyper-parameter that is determined by the final number of segments needed by the user. It can also be replaced by a distance threshold. Note that one and only one pair of point groups will be merged in each iteration.

\begin{algorithm}
\caption{Group merging.\label{alg:merge}}
\textbf{Input:} point cloud $S$, point group set $V$ with $K$ groups, number of iterations $T$. \\
\textbf{Output:} point group set $V$ with $K'$ groups

\begin{algorithmic}
\State $n \gets 0$
\While{$n < T$}

\ForAll{$G_i, G_j \in V$ and $i < j$}
   \State $D[G_{i}, G_{j}] \gets CID_g(G_i, G_j | S)$\Comment{\textit{cache $CID_g$}}
\EndFor
\State $G_{i^*}, {G_{j^*}} \gets \underset{i,j}{\argmin}(D[G_{i}, G_{j}])$
\State $G_{i^*} \gets G_{i^*} \cup G_{j^*}$
\Comment{\textit{Merge the two closest groups}}
\State $V \gets  V \setminus \{G_{j^*}\}$
\State $n \gets n+1$

\EndWhile
\end{algorithmic}
\end{algorithm}
\vspace{-5mm}
\label{sec:scene_abs}
\section{Experiments}
To demonstrate the effectiveness of CID we conduct several comprehensive experiments for the applications of CID explained in section~\ref{sec:app} on the  S3DIS~\cite{armeni_cvpr16} and ScanNet~\cite{dai2017scannet}.

\subsection{Point cloud instance segmentation via label propagation.}
\label{sec:label_prop_ex}
\textbf{Experiment setup.} \rebuttal{We randomly downsampled each scene to 20,000 points as} $S$. Then we use CID-FPS to propose $K=100$ seed points as $S_l$ for each point cloud and set $M=100$ for approximating CID. For these seed points, the ground truth semantic and instance labels are given. The labels are propagated to the 20,000 points via our approach. We then further propagate the labels of the 20,000 points to all of the remaining points based on the nearest neighbor in Euclidean space. To account for the randomness in CID-FPS, we run our experiment 5 times with 5 different random initial seed points. The reported performance is an average result. 

\textbf{Baseline methods.} \rebuttal{There have been many approaches for point cloud segmentation in recent years, such as~\cite{yi2019gspn, liang20203d}, and~\cite{chen2021lrgnet}}. We choose SGPN~\cite{wang2018sgpn} and PointGroup~\cite{jiang2020pointgroup} as the two baseline methods because SGPN is a classical instance segmentation neural network and PointGroup is a more recent instance segmentation neural network. 

\textbf{Evaluation.} We follow the same evaluation criteria as in SGPN~\cite{wang2018sgpn} and PointGroup~\cite{jiang2020pointgroup}. The IoU between each predicted instance segment and its corresponding ground truth instance segment is calculated. Then the average precision (AP) is calculated using three IoU thresholds ($\text{AP}_{25}$,  $\text{AP}_{50}$, $\text{AP}_{75}$) for our method. More details about AP can be found in~\cite{everingham2015pascal}. The evaluation is performed individually for each semantic category\footnote{The category-wise evaluation results for PointGroup are provided by the authors of that work.}, and the mean AP among all categories is also reported. Since our method does not follow the train-test setup as the learning-based methods, \textit{the evaluation for our method is over all scenes in the two datasets}, while the evaluation for the baseline methods is on their testing datasets. In the two baseline methods, only $\text{AP}_{50}$ was reported  for each individual semantic category. 

\begin{figure*}
    \vspace{-5mm}
    \parbox{7in}{\centering\includegraphics[width=0.175\textwidth]{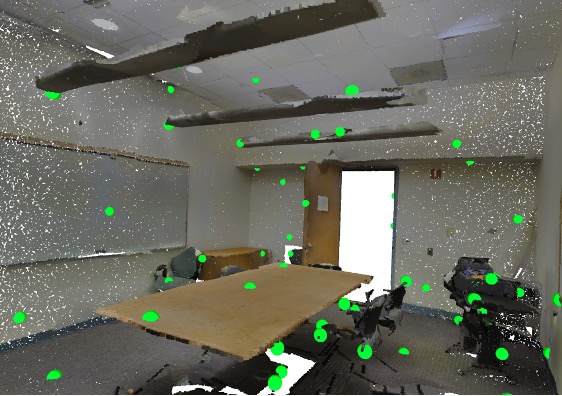}
    \includegraphics[width=0.175\textwidth]{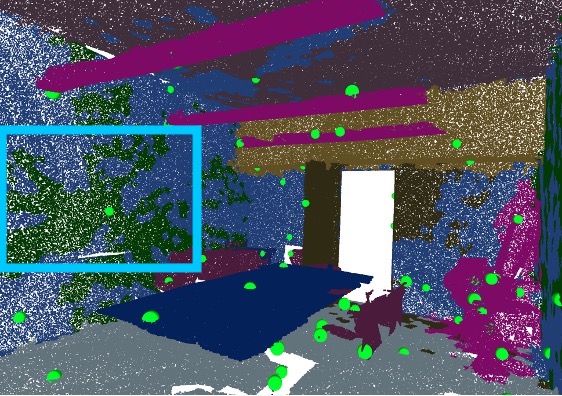}
    \includegraphics[width=0.175\textwidth]{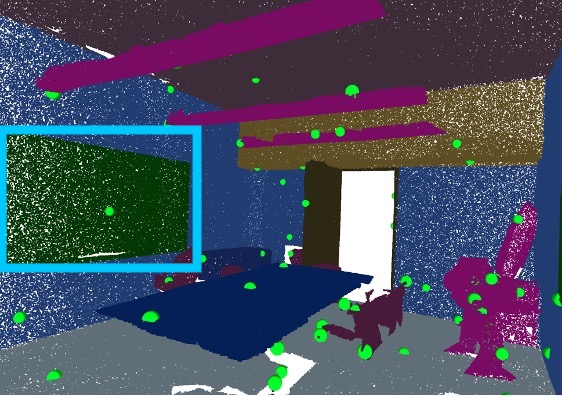}
    \includegraphics[width=0.175\textwidth]{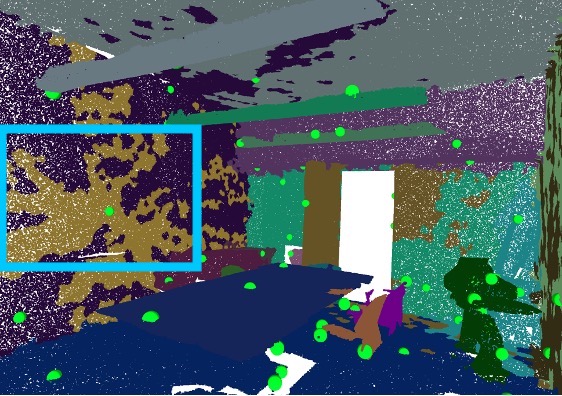}
    \includegraphics[width=0.175\textwidth]{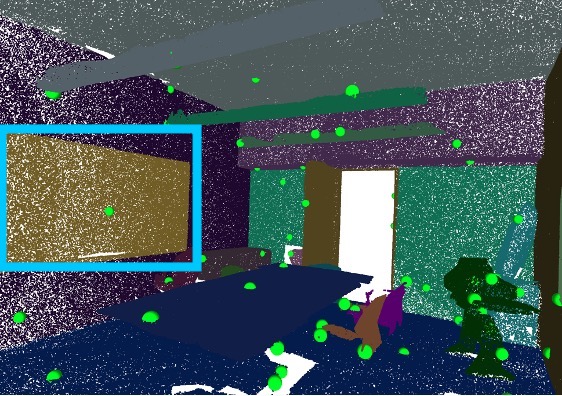}\\
    \includegraphics[width=0.175\textwidth]{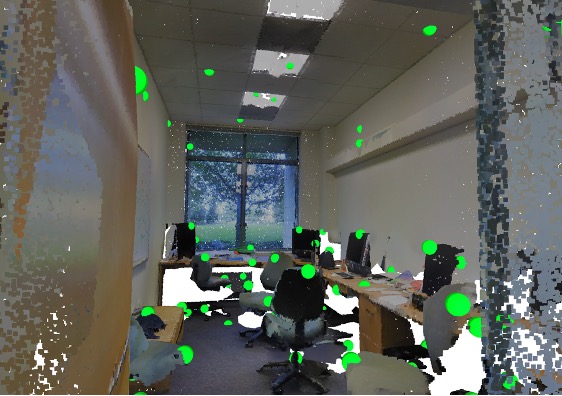}
    \includegraphics[width=0.175\textwidth]{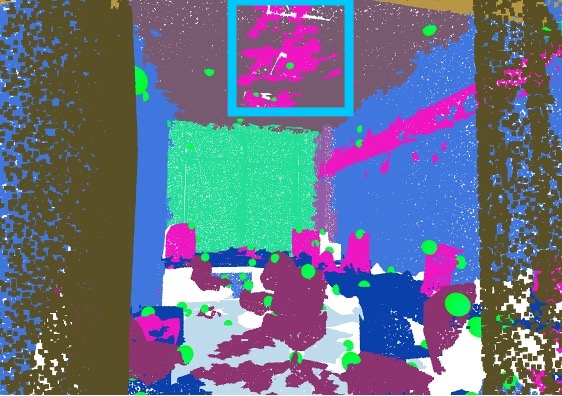}
    \includegraphics[width=0.175\textwidth]{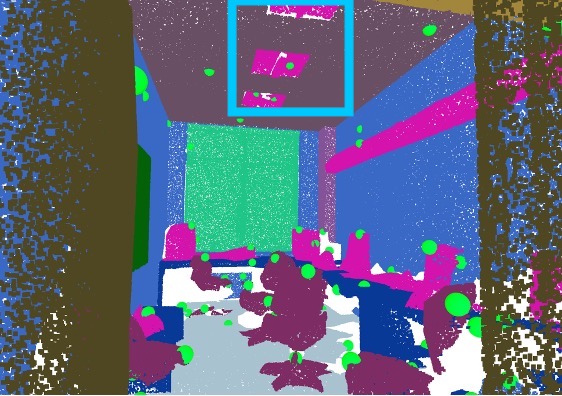}
    \includegraphics[width=0.175\textwidth]{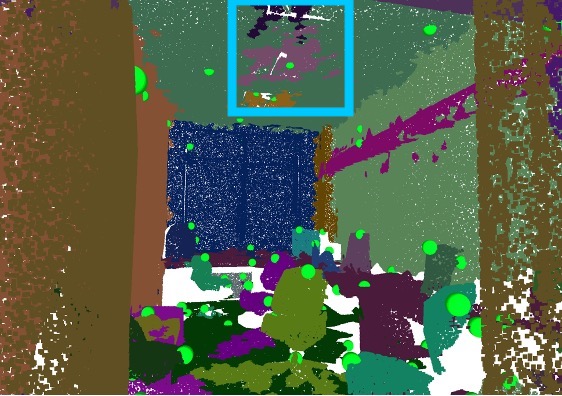}
    \includegraphics[width=0.175\textwidth]{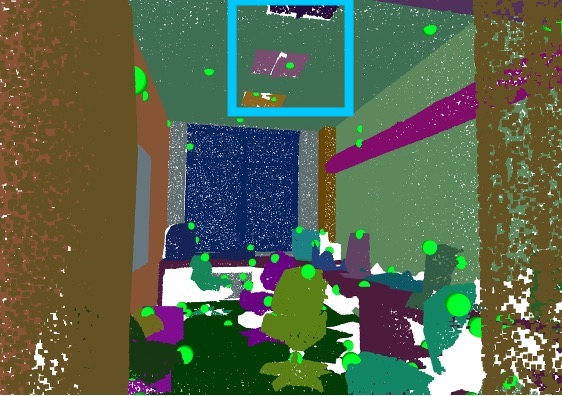}}
    \caption{\textbf{Qualitative results for CID-based point cloud instance segmentation.} From left to right: original point cloud, predicted semantic segmentation, ground truth semantic segmentation, predicted instance segmentation, and ground truth instance segmentation. Green balls indicate the locations of seed points. The boxes show the objects that cannot be well separated by the CID-based method, since they are nearly on the same plane (e.g., board and wall, ceiling and light).}
    \label{fig:label_prop}
\end{figure*}

\textbf{Results.}
Table~\ref{tab:label_prop_s3dis} and~\ref{tab:label_prop_scannet} shows the quantitative results of CID-based label propagation for instance segmentation. Note that, although our method \textit{does not} use color information of the point cloud that is used in both baseline methods, \textit{our method still outperforms the two baseline methods} in most of the semantic categories. The reason that our method has lower $\text{AP}$ on walls, doors, and boards is that these objects are sometimes overlapped with each other, which is difficult to be distinguished only using convexity without any color information (see Figure~\ref{fig:label_prop}).

\textbf{Discussion.} We want to emphasize that our purpose here is not to propose a new state-of-the-art instance segmentation approach. The comparison is not really apple-to-apple. Although our approach only needs a small portion of labeled points and does not require the training process, the seed points need to be proposed and labeled for any new point clouds, while the supervised learning approaches do not need labeling in the inference phase. The main purpose of this experiment is to show the potential of reducing manual labeling efforts via CID, such as an interactive labeling tool based on CID.

\begin{table*}[!t]
\caption{CID-based label propagation for instance segmentation on S3DIS~\cite{armeni_cvpr16}. 
\label{tab:label_prop_s3dis}}
\begin{center}
\begin{threeparttable}
\vspace{-8mm}
\resizebox{\textwidth}{!}{%
\begin{tabular}{
m{2.2cm} |  
>{\centering}m{0.8cm} 
>{\centering}m{0.8cm} 
>{\centering}m{0.8cm} 
>{\centering}m{0.8cm}
>{\centering}m{1.2cm}
>{\centering}m{1.2cm} 
>{\centering}m{0.8cm}
>{\centering}m{0.8cm}
>{\centering}m{0.8cm}
>{\centering}m{1.2cm}
>{\centering}m{0.8cm}
>{\centering}m{0.8cm}
>{\centering}m{0.8cm}
>{\centering}m{0.8cm}
>{\centering\arraybackslash}m{1.2cm} 
}
\toprule
 & Ceiling  & Floor & Wall & Beam  & Column & Window & Door
& Chair & Table & Bookcase & Board & Sofa & Stairs & Clutter
& \textbf{Mean} 
\\
\midrule
\multirow{3}{*}{Ours$^*$} & 0.956 & 0.995 & 0.903 & 0.979 & 0.955 & 0.963 & 0.768 & 0.877 & 0.917 & 0.926 & 0.833 & 0.958 & 0.883 & 0.842 & 0.911 \\ 
& \textbf{0.867} & \textbf{0.971} & 0.645 & \textbf{0.861} & \textbf{0.702} & \textbf{0.814} & 0.525 & 0.722 & \textbf{0.654} & \textbf{0.697} & 0.136 &\textbf{0.708} & \textbf{0.845} & \textbf{0.524} & \textbf{0.691} \\
& 0.639 & 0.780 & 0.300 & 0.539 & 0.147 & 0.248 & 0.176 & 0.402 & 0.234 & 0.260 & 0.024 & 0.361 & 0.568 & 0.180 & 0.347 \\
\midrule
SGPN~\cite{wang2018sgpn} & 0.794 & 0.663 & \textbf{0.888} & 0.780 & 0.607 & 0.666 & 0.568 & 0.408 & 0.470 & 0.476 & 0.111 & 0.064 & N/A & N/A & 0.541\\
\midrule
PointGroup~\cite{jiang2020pointgroup} & 0.724 & 0.966 & 0.454 & 0.627 & 0.393 & 0.808 & \textbf{0.593} & \textbf{0.887} & 0.567 & 0.431 & \textbf{0.785} & 0.565  &  N/A & 0.522 & 0.640\\
\bottomrule
\end{tabular}
}

\begin{tablenotes}
      \tiny
      \item $^*$For our approach, top/middle/bottom rows report $\text{AP}_{25}$/$\text{AP}_{50}$/$\text{AP}_{75}$. For the two baseline approaches, only $\text{AP}_{50}$ was reported by the original works.
    \end{tablenotes} 

\end{threeparttable}
\end{center}
\end{table*}

\begin{table*}[!t]
\vspace{-3mm}
\caption{CID-based label propagation for instance segmentation on ScanNet~\cite{dai2017scannet}. 
\label{tab:label_prop_scannet}}
\begin{center}
\begin{threeparttable}
\vspace{-8mm}
\resizebox{\textwidth}{!}{%
\begin{tabular}{
m{2cm} |  
>{\centering}m{0.8cm} 
>{\centering}m{0.8cm} 
>{\centering}m{0.8cm} 
>{\centering}m{0.8cm}
>{\centering}m{0.8cm}
>{\centering}m{0.8cm} 
>{\centering}m{0.8cm}
>{\centering}m{0.8cm}
>{\centering}m{0.8cm}
>{\centering}m{0.8cm}
>{\centering}m{0.8cm}
>{\centering}m{0.8cm}
>{\centering}m{0.8cm}
>{\centering}m{1.1cm}
>{\centering}m{0.8cm}
>{\centering}m{0.8cm}
>{\centering}m{0.8cm}
>{\centering}m{0.8cm}
>{\centering\arraybackslash}m{0.8cm} 
}
\toprule
 & cabinet & bed & chair & sofa & table & door & window & bookshe. & picture & counter & desk & curtain & refrige. & s. curtain & toilet & sink & bathtub & other
& \textbf{Mean} 
\\
\midrule
\multirow{3}{*}{Ours$^*$} & 0.911 & 0.989 & 0.942 & 1.0 & 0.941 & 0.89 & 0.896 & 0.911 & 0.568 & 0.701 & 0.835 & 0.952 & 0.945 & 0.993 & 0.992 & 0.837 & 0.986 & 0.868 & 0.898 \\ 
& \textbf{0.667} & \textbf{0.777} & 0.742 & 0.000 & \textbf{0.674} & \textbf{0.600} & \textbf{0.552} & 0.623 & 0.082 & \textbf{0.206} & 0.381 & \textbf{0.729} & \textbf{0.779} & 0.913 & 0.945 & 0.376 & 0.894 & \textbf{0.559} & 0.583
 \\
& 0.257 & 0.275 & 0.39 & 0.0 & 0.276 & 0.245 & 0.157 & 0.191 & 0.011 & 0.031 & 0.071 & 0.322 & 0.31 & 0.71 & 0.644 & 0.095 & 0.298 & 0.212 & 0.250 \\
\midrule
SGPN~\cite{wang2018sgpn} & 0.065 & 0.390 & 0.275 & 0.351 & 0.168 & 0.087 & 0.138 & 0.169 & 0.014 & 0.029 & 0.000 & 0.069 & 0.027 & 0.000 & 0.438 & 0.112 & 0.208 & 0.043 & 0.143\\
\midrule
PointGroup~\cite{jiang2020pointgroup} & 0.505 & 0.765 & \textbf{0.797} & \textbf{0.756} & 0.556 & 0.441 & 0.513 & \textbf{0.624} & \textbf{0.476} & 0.116 & \textbf{0.384} & 0.696 & 0.596 & \textbf{1.000} & \textbf{0.997} & \textbf{0.666} & \textbf{1.000} & \textbf{0.559} & \textbf{0.636}\\
\bottomrule
\end{tabular}
}

\begin{tablenotes}
      \tiny
      \item $^*$For our approach, top/middle/bottom rows report $\text{AP}_{25}$/$\text{AP}_{50}$/$\text{AP}_{75}$. For the two baseline approaches, only $\text{AP}_{50}$ was reported by the original works.
    \end{tablenotes} 

\end{threeparttable}
\end{center}
\end{table*}

\subsection{Convexity-based point cloud abstraction}

\begin{figure*}
    \vspace{-3mm}
    \parbox{7in}{\centering\includegraphics[width=0.15\textwidth]{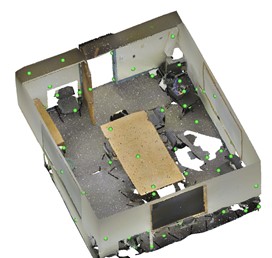}
    \includegraphics[width=0.15\textwidth]{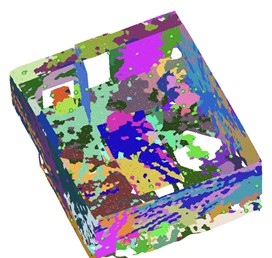}
    \includegraphics[width=0.15\textwidth]{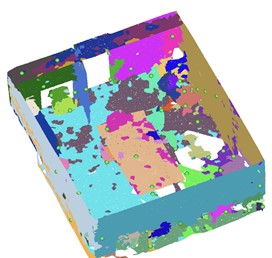}
    \includegraphics[width=0.15\textwidth]{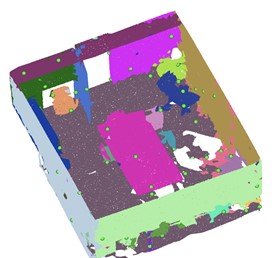}
    \includegraphics[width=0.15\textwidth]{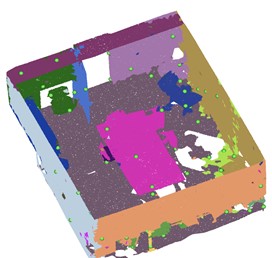}
    \includegraphics[width=0.15\textwidth]{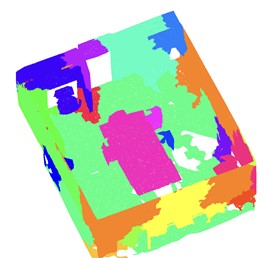} \\
    \includegraphics[width=0.14\textwidth]{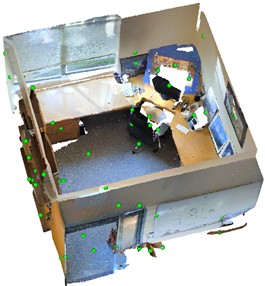}
    \includegraphics[width=0.14\textwidth]{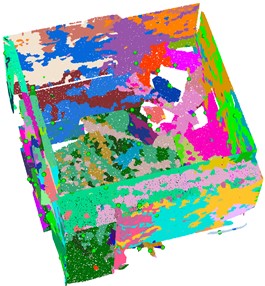}
    \includegraphics[width=0.14\textwidth]{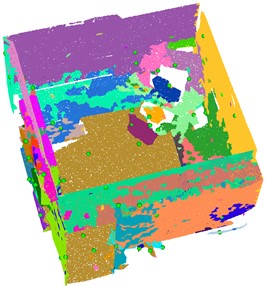}
    \includegraphics[width=0.14\textwidth]{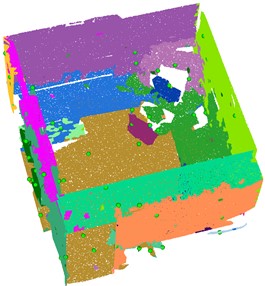}
    \includegraphics[width=0.14\textwidth]{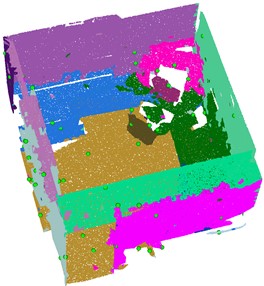}
    \includegraphics[width=0.14\textwidth]{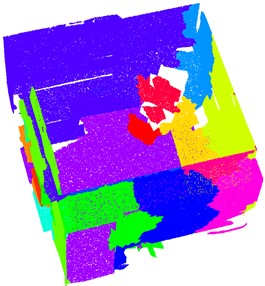}}
    \caption{\textbf{CID-based point cloud decomposition.} From left to right, the first column shows the original point cloud and the second to the fourth column shows our method at merging step $n=0, \ T_{1.0}/2, \ T_{1.0}$. The fifth column shows our method at merging step $n=T_b$, where our method shares the same compactness as the baseline method. The sixth column shows the baseline method~\cite{kaick2014shape}. Different colors represent different detected convex parts in a certain scene. Green dots indicate the locations of the seed points.}
    \label{fig:abs_quant}
    \vspace{-4mm}
\end{figure*}

\textbf{Experiment setup.} We also use S3DIS~\cite{armeni_cvpr16} dataset for this experiment, and the same settings as in~\ref{sec:label_prop_ex}. The only difference is that, in point cloud abstraction, there is no ground truth label. Therefore, the group indices are propagated to the whole point cloud from the 20,000 points instead of ground truth labels.

\textbf{Baseline methods.} To the best of our knowledge, existing convexity-based point cloud decomposition approaches cannot be directly applied to the unoriented point cloud.
Therefore, we choose~\cite{kaick2014shape}, which is the closest one to our approach, as the baseline approach.
One difference between the two approaches is that ours does not need the oriented normals while the baseline does. Therefore, before sending the point clouds to the baseline approach, we first estimate and orient the normals for the unoriented point clouds via the method provided by Open3D~\cite{Zhou2018}.

\textbf{Evaluation.} 
There could be different evaluation metrics for point cloud decomposition depending on different criteria. Considering scene understanding applications, the object instances should be preserved after the decomposition. Therefore, we define the following two evaluation metrics (in addition to the notations in~\ref{sec:scene_abs}, assume that there are $K_{gt}$ unique ground truth instance labels for the point cloud $S$; in each group, $G_i$, the number of majority points with the same ground truth instances label is $m_i$):

\begin{itemize}
    \item \textit{Compactness.} It is the ratio between the number of ground truth instances and the number of final groups:
    \begin{equation}
        \textit{Compactness}=K_{gt} / K'.
    \end{equation}
    \item \textit{Purity.} The purity is defined by the sum of the number of majority points with the same ground truth instance label over all groups divided by the total number of points in the point cloud:
    \begin{equation}
        \textit{Purity}=\frac{1}{|S|}\sum_{i=1}^{K'}m_i. 
    \end{equation}
\end{itemize}

With the increase in compactness, the segmentation is more concise, since the total number of segments is reduced. Note that the compactness can be larger than 1 when the number of groups output from the abstraction is smaller than the number of instances. The purity measures how much each group contains points from the same object instance. It is obvious that the maximum value of purity is 1 when each group only contains points from the same instance. Compactness and purity are usually inversely related. Higher compactness usually means lower purity, because, in a highly compact abstraction, points from different instances are more likely to be segmented into the same group. We compare the baseline method with our method at the same level of compactness for each individual scene, which is determined by the baseline method. We also plot the purity-compactness curve for our method on our project website.

\begin{figure}
    \centering
    \includegraphics[width=0.3\textwidth]{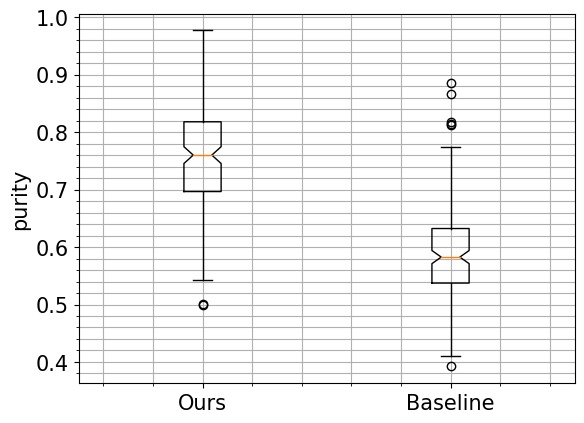}
    \caption{\textbf{Purity for scene abstraction.} Our method shows higher purity than the baseline method. The median purity of our method is 0.76, while the baseline method is 0.58.}
    \label{fig:purity}
    \vspace{-2mm}
\end{figure}

\textbf{Results.}
In Figure~\ref{fig:abs_quant}, we show the point cloud convex shape decomposition results at some critical $T$ values for group merging. $T_{1.0}$ is the value where the compactness reaches 1.0. $T_b$ is the value where the compactness reaches the same level as the baseline method. We can see that our method has a better ability in preserving the large convex parts (usually large objects), such as walls and floors. The better performance of our method is due to the independence from normal estimation and alignment, which may introduce errors to the baseline method. 
Figure~\ref{fig:purity} shows the boxplot of purity for our method and the baseline method \rebuttal{at the compactness level $T_b$}, where the median purity of our method exceeds the baseline method by a large margin (over 15\%). 

\begin{figure}
    \centering
    \parbox{3in}{\centering\includegraphics[width=0.13\textwidth]{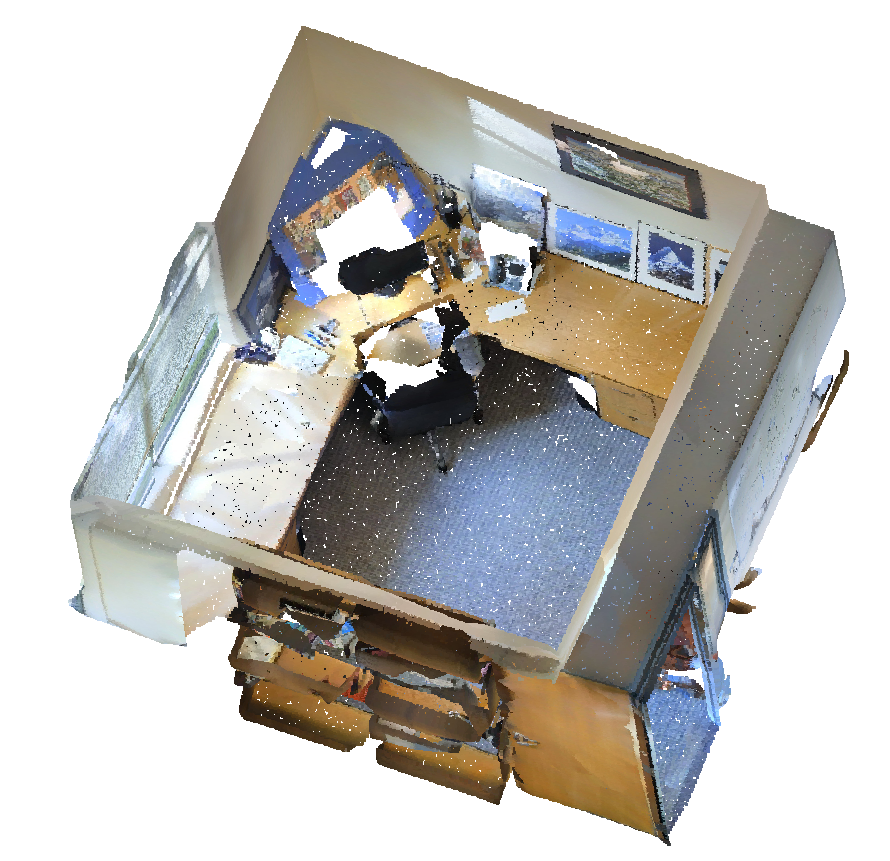}
    \includegraphics[width=0.13\textwidth]{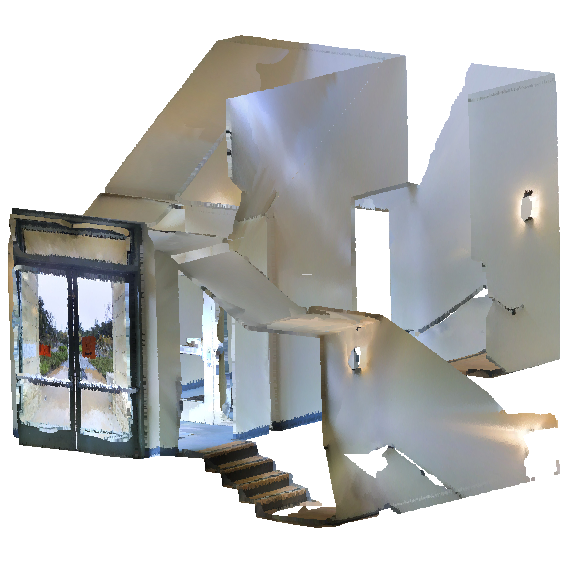}
    \includegraphics[width=0.13\textwidth]{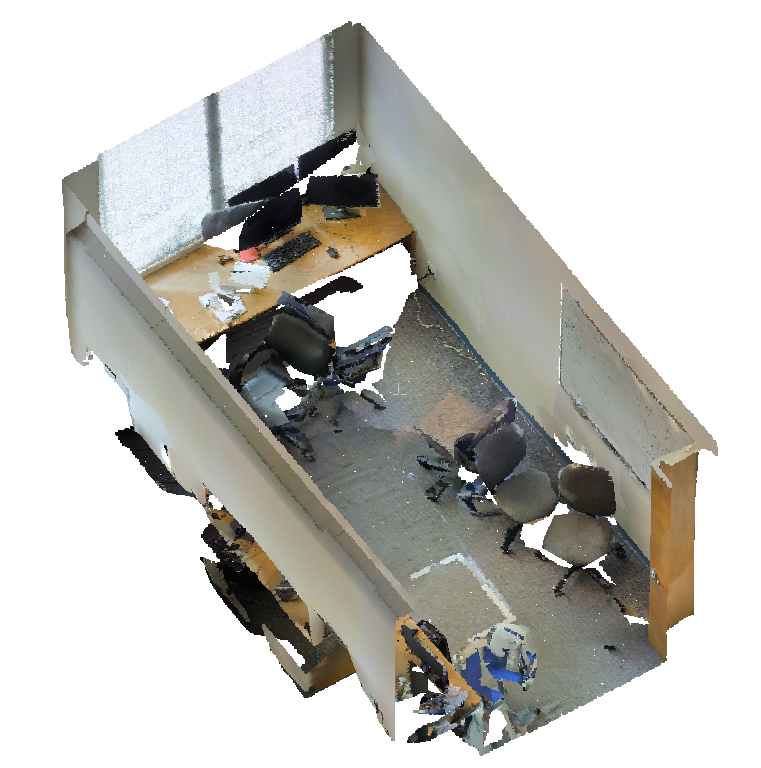}\\
    \includegraphics[width=0.13\textwidth]{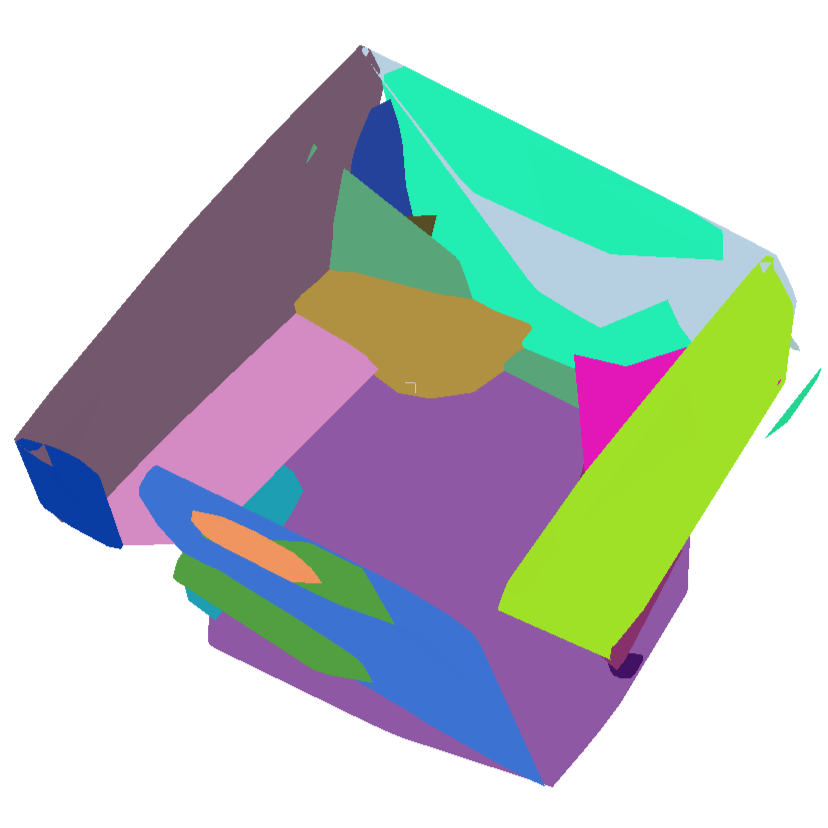}
    \includegraphics[width=0.13\textwidth]{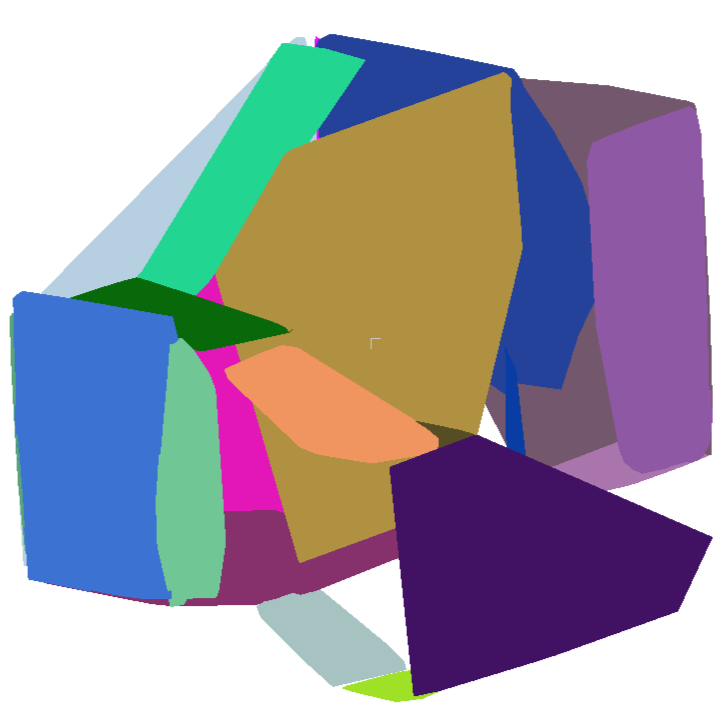}
    \includegraphics[width=0.13\textwidth]{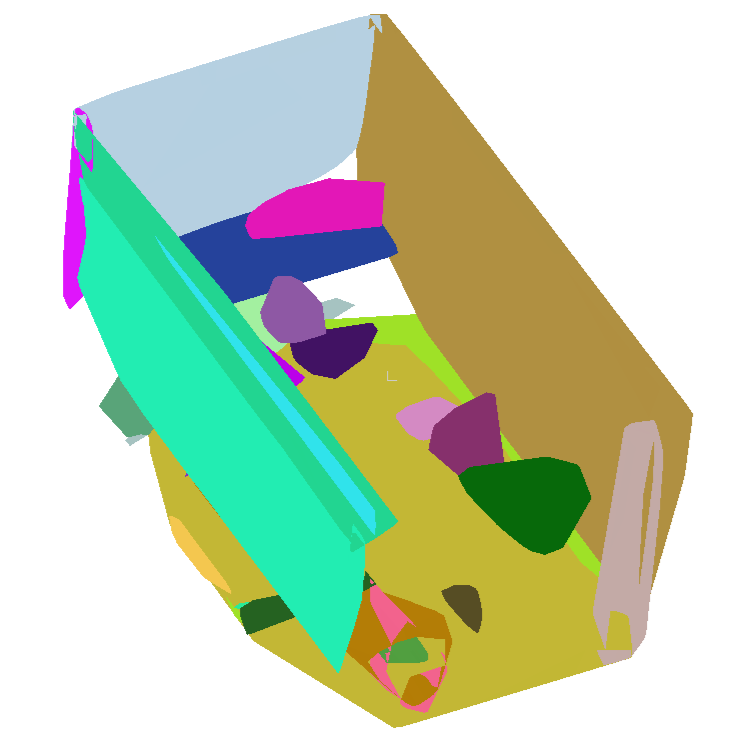}}
    \caption{\textbf{CID-based convex hull.} The first row shows the original point cloud, and the second row shows the corresponding convex hull computed by our method. Different colors represent different convex hull parts.}
    \label{fig:convex_hull}  
    \vspace{-5mm}
\end{figure}
\section{Algorithm Analysis}
Next, we perform further experiments on our method to show how the number of seed points can influence our method. We will also discuss the limitations. On our project website, we will perform a robustness analysis on CID, regarding the noise in point cloud.

\begin{figure}
    \centering
    \includegraphics[width=0.4\textwidth]{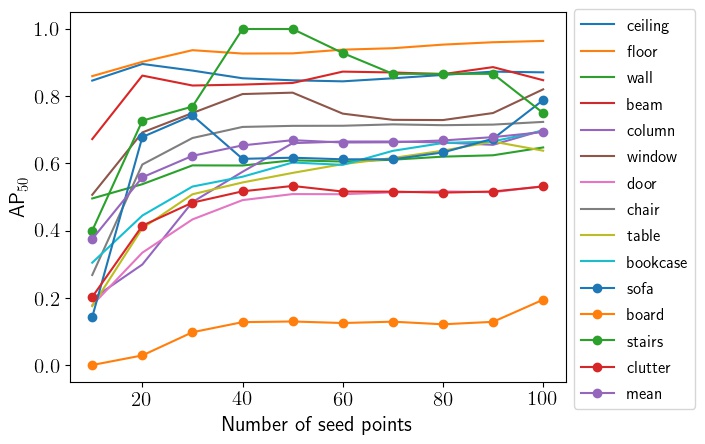}
    \vspace{-2mm}
    \caption{The effects of \#seed-points on $\text{AP}_{50}$ for various objects.}
    \label{fig:num_seed}
    \vspace{-5mm}
\end{figure}

\textbf{Effects on the number of seed points.} We evaluate the CID-based label propagation using the different number of seed points. The number of seed points proposed by CID-FPS is incrementally increased from 10 to 100 with a step of 10. Figure~\ref{fig:num_seed} shows the change of mean $\text{AP}_{50}$ with increasing the number of seed points. 

We can see that with the increasing number of seed points, the $\text{AP}_{50}$ for most of the classes of object increases, and finally saturates after a certain number of seed points. This is because when the number of seed points is small, some objects are not covered by any seed points. Therefore, it is impossible to correctly segment these objects. When the number of seed points reaches a certain level, most of the convex parts in the scene contain at least one seed point. In this case, increasing the number of seed points further has very limited influence.

Interestingly, we find that some classes of objects are less influenced by the number of seed points, such as the ceiling, floor, and beam. The explanation of such a phenomenon is that our CID-FPS seed point proposal method tends to first find seed points that have larger CID from other points in the point cloud. In indoor scenes, ceilings, floors, and beams usually have higher concavity from other objects compared with smaller objects in the scene, such as chairs and tables. Therefore, with a small number of seed points, ceilings, floors, and beams still have a higher chance to be sampled by CID-FPS. 

Another interesting phenomenon is that the $\text{AP}_{50}$ for stairs first increases and then decreases. This is due to the fact that in the whole dataset, there are only 14 stair instances. At first, adding the number of seed points increases the chance that the stairs can be correctly segmented. With the increasing number of seed points, some seed points that are not on the stairs may propagate their labels to the points on the stairs, which causes a decrease in $\text{AP}_{50}$. Due to the very less number of instances (14), the variance of $\text{AP}_{50}$ is large even if only a small number of objects is influenced.

\textbf{Limitations of our method.} There are several limitations for CID in point cloud analysis:
\begin{itemize}
    \item \textit{Computational cost.} Since the CID-FPS process is iterative, the computational cost of CID-FPS can be large. The state-of-the-art supervised learning methods usually have inference time that is lower than 1 second per scene, while our label propagation method takes over 1 minute to process 1 million points (slow mainly due to the CID-FPS computation). 
    \item \textit{Non-uniformly sampled point clouds}. Another limitation of CID is that it currently cannot effectively handle outdoor LiDAR point clouds that are sparse and not uniformly sampled. Therefore we have to focus on indoor point clouds that are typically densely and uniformly sampled everywhere on indoor object surfaces.
    \item \textit{Thin object separation}. Another minor limitation of CID is that it cannot be used as the only cue to segment thin objects such as papers on the desk or paintings on the wall, as shown in previous experiments. However, together with visual features, this could be overcome.
\end{itemize}

\section{Conclusion}
Our proposed Concavity-Induced Distance (CID), which is the first distance that can measure the concavity between two points or two sets of points on an unoriented point cloud, has shown strong potential in indoor point cloud understanding tasks, such as instance segmentation and convexity-based point cloud segmentation, without heavy manual labeling as required by supervised learning methods. Our future work will focus on extending CID into more point cloud-based tasks such as object detection and outdoor scenes.

\bibliographystyle{ieeetran}
\bibliography{ref}

\begin{thebibliography}{10}
\providecommand{\url}[1]{#1}
\csname url@rmstyle\endcsname
\providecommand{\newblock}{\relax}
\providecommand{\bibinfo}[2]{#2}
\providecommand\BIBentrySTDinterwordspacing{\spaceskip=0pt\relax}
\providecommand\BIBentryALTinterwordstretchfactor{4}
\providecommand\BIBentryALTinterwordspacing{\spaceskip=\fontdimen2\font plus
\BIBentryALTinterwordstretchfactor\fontdimen3\font minus
  \fontdimen4\font\relax}
\providecommand\BIBforeignlanguage[2]{{%
\expandafter\ifx\csname l@#1\endcsname\relax
\typeout{** WARNING: IEEEtran.bst: No hyphenation pattern has been}%
\typeout{** loaded for the language `#1'. Using the pattern for}%
\typeout{** the default language instead.}%
\else
\language=\csname l@#1\endcsname
\fi
#2}}

\bibitem{schulman2013finding}
J.~Schulman, J.~Ho, A.~X. Lee, I.~Awwal, H.~Bradlow, and P.~Abbeel, ``Finding
  locally optimal, collision-free trajectories with sequential convex
  optimization.'' in \emph{Robotics: science and systems}, vol.~9, no.~1.\hskip
  1em plus 0.5em minus 0.4em\relax Berlin, Germany, 2013, pp. 1--10.

\bibitem{stein2014convexity}
S.~C. Stein, F.~W{\"o}rg{\"o}tter, M.~Schoeler, J.~Papon, and T.~Kulvicius,
  ``Convexity based object partitioning for robot applications,'' in
  \emph{ICRA}.\hskip 1em plus 0.5em minus 0.4em\relax IEEE, 2014, pp.
  3213--3220.

\bibitem{chari2012convex}
V.~Chari, A.~Agrawal, Y.~Taguchi, and S.~Ramalingam, ``Convex bricks: A new
  primitive for visual hull modeling and reconstruction,'' in
  \emph{ICRA}.\hskip 1em plus 0.5em minus 0.4em\relax IEEE, 2012, pp. 770--777.

\bibitem{qin2014real}
S.~Qin, X.~Zhu, Y.~Yang, and Y.~Jiang, ``Real-time hand gesture recognition
  from depth images using convex shape decomposition method,'' \emph{Journal of
  Signal Processing Systems}, vol.~74, no.~1, pp. 47--58, 2014.

\bibitem{ghosh2013fast}
M.~Ghosh, N.~M. Amato, Y.~Lu, and J.-M. Lien, ``Fast approximate convex
  decomposition using relative concavity,'' \emph{Computer-Aided Design},
  vol.~45, no.~2, pp. 494--504, 2013.

\bibitem{mamou2016volumetric}
K.~Mamou, E.~Lengyel, and A.~Peters, ``Volumetric hierarchical approximate
  convex decomposition,'' in \emph{Game Engine Gems 3}.\hskip 1em plus 0.5em
  minus 0.4em\relax AK Peters, 2016, pp. 141--158.

\bibitem{lien2004approximate}
J.-M. Lien and N.~M. Amato, ``Approximate convex decomposition of polygons,''
  in \emph{Proceedings of the twentieth annual symposium on Computational
  geometry}, 2004, pp. 17--26.

\bibitem{asafi2013weak}
S.~Asafi, A.~Goren, and D.~Cohen-Or, ``Weak convex decomposition by
  lines-of-sight,'' in \emph{Computer graphics forum}, vol.~32, no.~5.\hskip
  1em plus 0.5em minus 0.4em\relax Wiley Online Library, 2013, pp. 23--31.

\bibitem{kaick2014shape}
O.~V. Kaick, N.~Fish, Y.~Kleiman, S.~Asafi, and D.~Cohen-Or, ``Shape
  segmentation by approximate convexity analysis,'' \emph{ACM Transactions on
  Graphics (TOG)}, vol.~34, no.~1, pp. 1--11, 2014.

\bibitem{christoph2014object}
S.~Christoph~Stein, M.~Schoeler, J.~Papon, and F.~Worgotter, ``Object
  partitioning using local convexity,'' in \emph{Proceedings of CVPR}, 2014,
  pp. 304--311.

\bibitem{gong2017point}
X.~Gong, M.~Chen, and X.~Yang, ``Point cloud segmentation of 3d scattered parts
  sampled by realsense,'' in \emph{2017 IEEE International Conference on
  Information and Automation (ICIA)}.\hskip 1em plus 0.5em minus 0.4em\relax
  IEEE, 2017, pp. 1--6.

\bibitem{2017arXiv170201105A}
I.~{Armeni}, A.~{Sax}, A.~R. {Zamir}, and S.~{Savarese}, ``{Joint
  2D-3D-Semantic Data for Indoor Scene Understanding},'' \emph{ArXiv e-prints},
  Feb. 2017.

\bibitem{kreavoy2007model}
V.~Kreavoy, D.~Julius, and A.~Sheffer, ``Model composition from interchangeable
  components,'' in \emph{15th Pacific Conference on Computer Graphics and
  Applications (PG'07)}.\hskip 1em plus 0.5em minus 0.4em\relax IEEE, 2007, pp.
  129--138.

\bibitem{lien2007approximate}
J.~Lien and N.~M. Amato, ``Approximate convex decomposition of polyhedra,'' in
  \emph{Proceedings of the 2007 ACM symposium on Solid and physical modeling},
  2007, pp. 121--131.

\bibitem{lien2008approximate}
J.~M. Lien and N.~M. Amato, ``Approximate convex decomposition of polyhedra and
  its applications,'' \emph{Computer Aided Geometric Design}, vol.~25, no.~7,
  pp. 503--522, 2008.

\bibitem{sheffer2007shuffler}
V.~K. D. J.~A. Sheffer, ``Shuffler: Modeling with interchangeable parts,''
  \emph{Visual Computer journal}, 2007.

\bibitem{liu2010convex}
H.~Liu, W.~Liu, and L.~J. Latecki, ``Convex shape decomposition,'' in
  \emph{CVPR}.\hskip 1em plus 0.5em minus 0.4em\relax IEEE, 2010, pp. 97--104.

\bibitem{shinagawa1991surface}
Y.~Shinagawa, T.~L. Kunii, and Y.~L. Kergosien, ``Surface coding based on morse
  theory,'' \emph{IEEE Computer Graphics and Applications}, vol.~11, no.~05,
  pp. 66--78, 1991.

\bibitem{tateno2015real}
K.~Tateno, F.~Tombari, and N.~Navab, ``Real-time and scalable incremental
  segmentation on dense slam,'' in \emph{IROS}.\hskip 1em plus 0.5em minus
  0.4em\relax IEEE, 2015, pp. 4465--4472.

\bibitem{deng2020cvxnet}
B.~Deng, K.~Genova, S.~Yazdani, S.~Bouaziz, G.~Hinton, and A.~Tagliasacchi,
  ``Cvxnet: Learnable convex decomposition,'' in \emph{CVPR}, 2020, pp. 31--44.

\bibitem{gadelha2020label}
M.~Gadelha, A.~RoyChowdhury, G.~Sharma, E.~Kalogerakis, L.~Cao,
  E.~Learned-Miller, R.~Wang, and S.~Maji, ``Label-efficient learning on point
  clouds using approximate convex decompositions,'' in \emph{ECCV}.\hskip 1em
  plus 0.5em minus 0.4em\relax Springer, 2020, pp. 473--491.

\bibitem{tulsiani2017learning}
S.~Tulsiani, H.~Su, L.~J. Guibas, A.~A. Efros, and J.~Malik, ``Learning shape
  abstractions by assembling volumetric primitives,'' in \emph{CVPR}, 2017, pp.
  2635--2643.

\bibitem{zou20173d}
C.~Zou, E.~Yumer, J.~Yang, D.~Ceylan, and D.~Hoiem, ``3d-prnn: Generating shape
  primitives with recurrent neural networks,'' in \emph{ICCV}, 2017, pp.
  900--909.

\bibitem{qi2017pointnet++}
C.~R. Qi, L.~Yi, H.~Su, and L.~J. Guibas, ``Pointnet++: Deep hierarchical
  feature learning on point sets in a metric space,'' \emph{arXiv preprint
  arXiv:1706.02413}, 2017.

\bibitem{feng2014fast}
C.~Feng, Y.~Taguchi, and V.~R. Kamat, ``Fast plane extraction in organized
  point clouds using agglomerative hierarchical clustering,'' in
  \emph{ICRA}.\hskip 1em plus 0.5em minus 0.4em\relax IEEE, 2014, pp.
  6218--6225.

\bibitem{armeni_cvpr16}
I.~Armeni, O.~Sener, A.~R. Zamir, H.~Jiang, I.~Brilakis, M.~Fischer, and
  S.~Savarese, ``3d semantic parsing of large-scale indoor spaces,'' in
  \emph{CVPR}, 2016.

\bibitem{dai2017scannet}
A.~Dai, A.~X. Chang, M.~Savva, M.~Halber, T.~Funkhouser, and M.~Nie{\ss}ner,
  ``Scannet: Richly-annotated 3d reconstructions of indoor scenes,'' in
  \emph{Proc. CVPR}, 2017.

\bibitem{yi2019gspn}
L.~Yi, W.~Zhao, H.~Wang, M.~Sung, and L.~J. Guibas, ``Gspn: Generative shape
  proposal network for 3d instance segmentation in point cloud,'' in
  \emph{CVPR}, 2019, pp. 3947--3956.

\bibitem{liang20203d}
Z.~Liang, M.~Yang, H.~Li, and C.~Wang, ``3d instance embedding learning with a
  structure-aware loss function for point cloud segmentation,'' \emph{IEEE
  Robotics and Automation Letters}, vol.~5, no.~3, pp. 4915--4922, 2020.

\bibitem{chen2021lrgnet}
J.~Chen, Z.~Kira, and Y.~K. Cho, ``Lrgnet: Learnable region growing for
  class-agnostic point cloud segmentation,'' \emph{IEEE Robotics and Automation
  Letters}, vol.~6, no.~2, pp. 2799--2806, 2021.

\bibitem{wang2018sgpn}
W.~Wang, R.~Yu, Q.~Huang, and U.~Neumann, ``Sgpn: Similarity group proposal
  network for 3d point cloud instance segmentation,'' in \emph{CVPR}, 2018, pp.
  2569--2578.

\bibitem{jiang2020pointgroup}
L.~Jiang, H.~Zhao, S.~Shi, S.~Liu, C.-W. Fu, and J.~Jia, ``Pointgroup: Dual-set
  point grouping for 3d instance segmentation,'' in \emph{CVPR}, 2020, pp.
  4867--4876.

\bibitem{everingham2015pascal}
M.~Everingham, S.~A. Eslami, L.~Van~Gool, C.~K. Williams, J.~Winn, and
  A.~Zisserman, ``The pascal visual object classes challenge: A
  retrospective,'' \emph{IJCV}, vol. 111, no.~1, pp. 98--136, 2015.

\bibitem{Zhou2018}
Q.-Y. Zhou, J.~Park, and V.~Koltun, ``{Open3D}: {A} modern library for {3D}
  data processing,'' \emph{arXiv:1801.09847}, 2018.

\end{thebibliography}
\end{document}